%% file: Main.tex
\documentclass[letterpaper]{article} 
\usepackage{aaai25}  
\usepackage{times}  
\usepackage{helvet}  
\usepackage{courier}  
\usepackage[hyphens]{url}  
\usepackage{graphicx} 
\usepackage{natbib}  
\usepackage{caption} 
\usepackage{algorithm}
\usepackage{algorithmic}

\usepackage{newfloat}
\usepackage{listings}
\DeclareCaptionStyle{ruled}{labelfont=normalfont,labelsep=colon,strut=off} 
\floatstyle{ruled}
\newfloat{listing}{tb}{lst}{}
\floatname{listing}{Listing}
\usepackage{amsthm,amsmath,amssymb}
\usepackage{mathrsfs}
\usepackage{bm}
\usepackage{booktabs}
\usepackage{multirow}
\usepackage{makecell}
\usepackage{colortbl}
\usepackage{xcolor}
\usepackage{array}
\usepackage{pifont}
\usepackage{tabularx}
\usepackage{cite}
\usepackage{tikz}
\usepackage{subfigure}
\usepackage{textcomp}

\title{Text Proxy: Decomposing Retrieval from a 1-to-$N$ Relationship \\ into $N$ 1-to-1 Relationships for Text-Video Retrieval}
\author{
    Jian Xiao,
    Zhenzhen Hu\thanks{Corresponding author.},
    Jia Li,
    Richang Hong
}
\affiliations{

    School of Computer Science and Information Engineering, Hefei University of Technology, Hefei, China\\
    j.xiao\_hfut@mail.hfut.edu.cn, \{zzhu, lijia\}@hfut.edu.cn, hongrc.hfut@gmail.com
}

\usepackage{bibentry}

\begin{document}

\maketitle

\begin{abstract}
Text-video retrieval (TVR) has seen substantial advancements in recent years, fueled by the utilization of pre-trained models and large language models (LLMs). Despite these advancements, achieving accurate matching in TVR remains challenging due to inherent disparities between video and textual modalities and irregularities in data representation. In this paper, we propose Text-Video-ProxyNet (TV-ProxyNet), a novel framework designed to decompose the conventional 1-to-$N$ relationship of TVR into $N$ distinct 1-to-1 relationships. By replacing a single text query with a series of text proxies, TV-ProxyNet not only broadens the query scope but also achieves a more precise expansion. Each text proxy is crafted through a refined iterative process, controlled by mechanisms we term as the director and dash, which regulate the proxy's direction and distance relative to the original text query. This setup not only facilitates more precise semantic alignment but also effectively manages the disparities and noise inherent in multimodal data. Our experiments on three representative video-text retrieval benchmarks, MSRVTT, DiDeMo, and ActivityNet Captions, demonstrate the effectiveness of TV-ProxyNet. The results show an improvement of 2.0\% to 3.3\% in R@1 over the baseline. TV-ProxyNet achieved state-of-the-art performance on MSRVTT and ActivityNet Captions, and a 2.0\% improvement on DiDeMo compared to existing methods, validating our approach's ability to enhance semantic mapping and reduce error propensity.
\end{abstract}

\begin{links}
    \link{Code}{https://github.com/musicman217/Text-Proxy/tree/main}
\end{links}

\section{Introduction}

Text-video retrieval~(TVR) is a crucial task of video understanding and  has garnered substantial attention in recent years~\cite{liu2021hit, ye2023hitea, jin2023text, wang2024text}. As a fundamentally 1-to-$N$ matching problem, the goal of TVR is to accurately link a text query to its corresponding video segment. The inherent disparity between video and textual modalities presents significant challenges for this task.

Achieving precise alignment demands sophisticated models capable of understanding and correlating complex multimodal data. Recent breakthroughs have been driven by leveraging pre-trained models~\cite{liu2021swin, fu2021violet, wang2023all, cheng2023vindlu} and large language models (LLMs)~\cite{wu2023cap4video,wang2024havtr}. These models, leveraging extensive training datasets and substantial computational power, have achieved significant advancements in cross-modal representation and alignment and have brought significant enhancements to the performance of TVR.

\begin{figure}[!tbp]
    \centering

    \begin{tikzpicture}[baseline]
        \node[anchor=west] (title1) at (0,0) {\small{(a)}};
        \node[anchor=west] (image1) at (title1.east) [xshift=1em] {\includegraphics[width=0.8\linewidth]{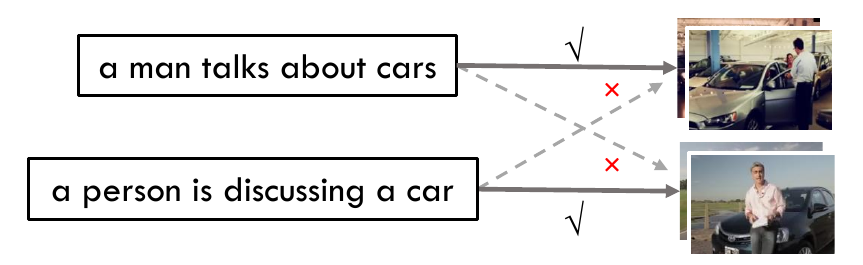}};
    \end{tikzpicture}

    \par
    \begin{tikzpicture}[baseline]
        \node[anchor=west] (title2) at (0,0) {\small{(b)}};
        \node[anchor=west] (image2) at (title2.east) [xshift=1em] {\includegraphics[width=0.8\linewidth]{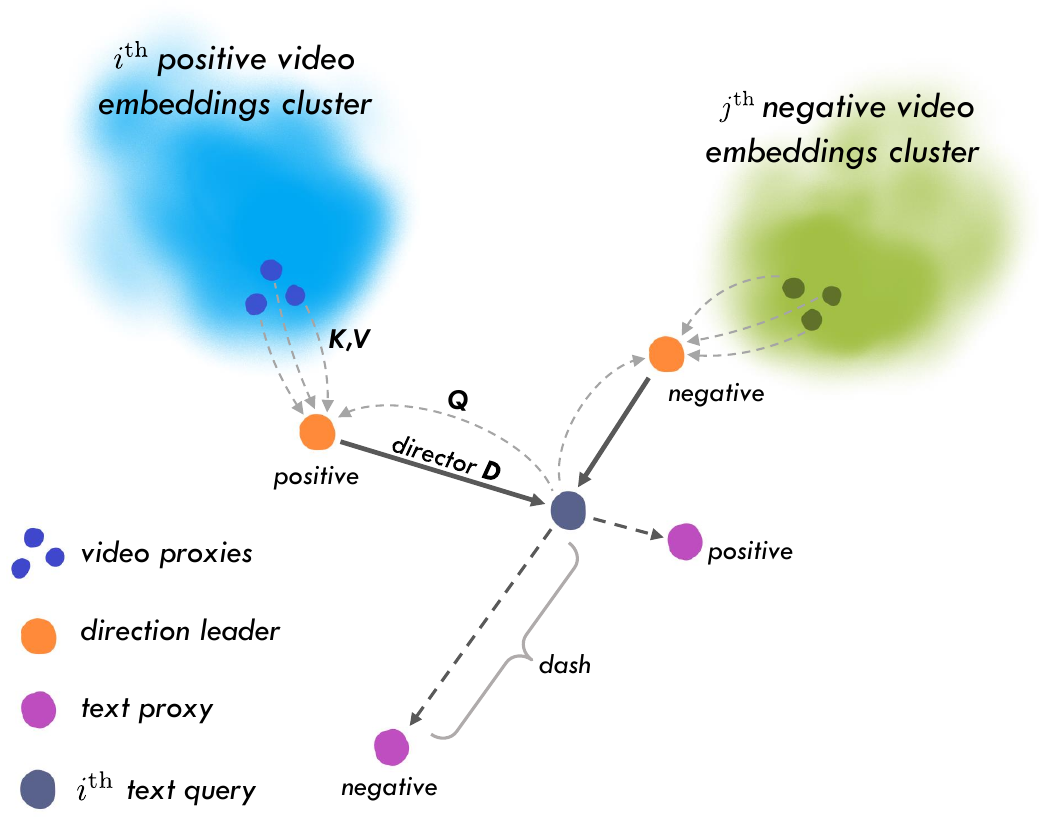}};
    \end{tikzpicture}

    \caption{(a): Text annotations being overly simplistic and providing insufficient distinctiveness for video is one of the irregularity problems. (b): Illustration of text proxy. The director leader and the dash mechanism influence the direction and distance of proxies, effectively drawing them closer to positive and further from negative video token clusters. The concept of video proxies is from the CLIP-ViP~\cite{xue2023clip}, and different video proxies can be seen as different video features containing different semantics.}
    \label{fig:1}
\end{figure}

\begin{figure*}[t]
    \centering
    \includegraphics[width=\textwidth]{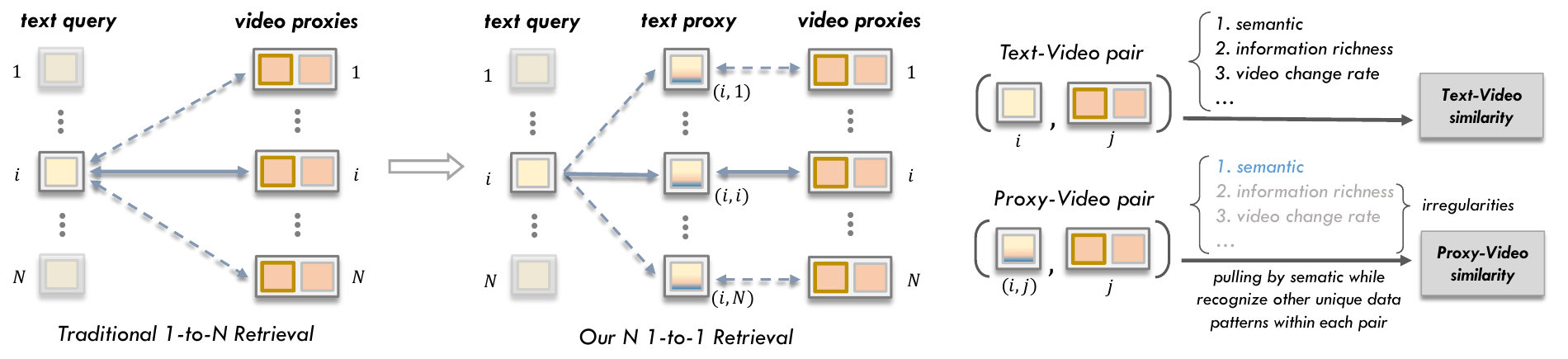}
    \caption{TV-ProxyNet converts the traditional 1-to-$N$ relationship into $N$ distinct 1-to-1 relationships. Each 1-to-1 relationship is unique, compelling the model to recognize the unique data patterns inherent in each pair, excluding irregular factors. }
    \label{fig:2}
\end{figure*}

However, accurate matching in TVR tasks requires more than just powerful representations. There are also inherent data issues~\cite{lin2022text,wang2022multi,wang2024text,wang2024havtr}. For instance, some text annotations are overly concise and lack the necessary distinctiveness between different videos.  Meanwhile, some videos exhibit complex content changes, corresponding to different texts at various stages, as shown in Fig.~\ref{fig:1}(a). These irregularities can lead to a paradox where models trained in other dataset spaces with more precise semantic alignment are more prone to errors in TVR matching. Several research efforts have already been dedicated to expanding the matching scope of a single query. 
By transforming a single text query into a multi-text query, \cite{wang2022multi} broaden the semantic scope of the query, reframing the task as a $K$-to-$N$ matching.
\cite{wang2024text} start from the feature space of text and treat text as a stochastic embedding to enrich the embedding, transforming the task into a space-to-$N$ matching. Although these methods expand the query scope by enriching the text, they lack specific associations with videos during expansion, missing clear directions and quantifiable values for the enrichment.

Our observations highlight that: the matching algorithms are predominantly based on semantics, neglecting other mighty factors that introduce noise and lack consistent patterns across various samples. These factors include, but are not limited to: disparate levels of information richness between videos and texts, and videos that range from having overly smooth frames to those with abrupt changes, which are not consistently reflected in the associated texts. Such discrepancies lead the confusion during the alignment process, as the model struggles to discern patterns amid these irregularities. We term this challenge the \textit{``Modeling Irregularity''} problem, where the model erroneously attempts to extract discernible patterns from semantic inconsistencies.

To alleviate the Modeling Irregularity problem, we propose a framework named \textbf{T}ext-\textbf{V}ideo-ProxyNet (TV-ProxyNet). This strategic replaces the single text query with a series of \textbf{text proxies} and effectively converts the typical 1-to-$N$ relationship of TVR into $N$ distinct 1-to-1 relationships. Each 1-to-1 relationship is unique, compelling the model to recognize and adapt to the unique data patterns inherent in each pair, not other irregular factors, as shown in Fig.~\ref{fig:2}. 
In generating the text proxy, TV-ProxyNet iteratively refines the proxy to balance the textual and video information effectively. 
To regulate the distribution of proxies, we control the direction and distance of each proxy through the coordinated efforts of the director and dash mechanisms. The director, in particular, is a directional vector determined by the direction leader and the query.
Directed by specific cues from a director, each text proxy leverages a dash to quantitatively adjust the proximity between positive and negative samples, effectively narrowing the gap with positive samples while distancing from negative ones. With each iteration, the direction and distance  to adjust the proxy relative to the original text query is assessed and the video information proportion increases. This iterative process is termed as the text proxy path.  Fig.~\ref{fig:1}(b) intuitively illustrates the interactive relationships between the text proxy, the query, the dash, and the director. This approach not only streamlines the retrieval process but also enhances the precision with which semantic relationships are mapped, significantly reducing the propensity for error introduced by irregular factors. Experiments on three representative video-text retrieval benchmarks demonstrate the effectiveness of TV-ProxyNet and provide new insights for building better retrieval models. The contributions of this paper are threefold:
\begin{itemize}
    \item We disentangle TVR from a 1-to-$N$ relationship into $N$ 1-to-1 relationships to alleviate the impact of irregular factors across numerous sample pairs.
    \item We propose a novel framework, TV-ProxyNet, to implement this $N$ 1-to-1 relationships, where the direction and distance of text proxy relative to the query is learnable.
    \item Extensive experiments on MSRVTT, DiDeMo and ActivityNet Captions show promising performance on various metrics while outperforms state-of-the-art methods.
\end{itemize}

\section{Related Work}

\subsection{Text-Video Retrieval}
Existing works can be divided into two categories: one is that using pre-trained video and text extractors to obtain offline features. Another is end-to-end training methods mostly benefiting from CLIP \cite{radford2021learning} model.

\begin{figure*}[t]
    \centering
    \includegraphics[width=\textwidth]{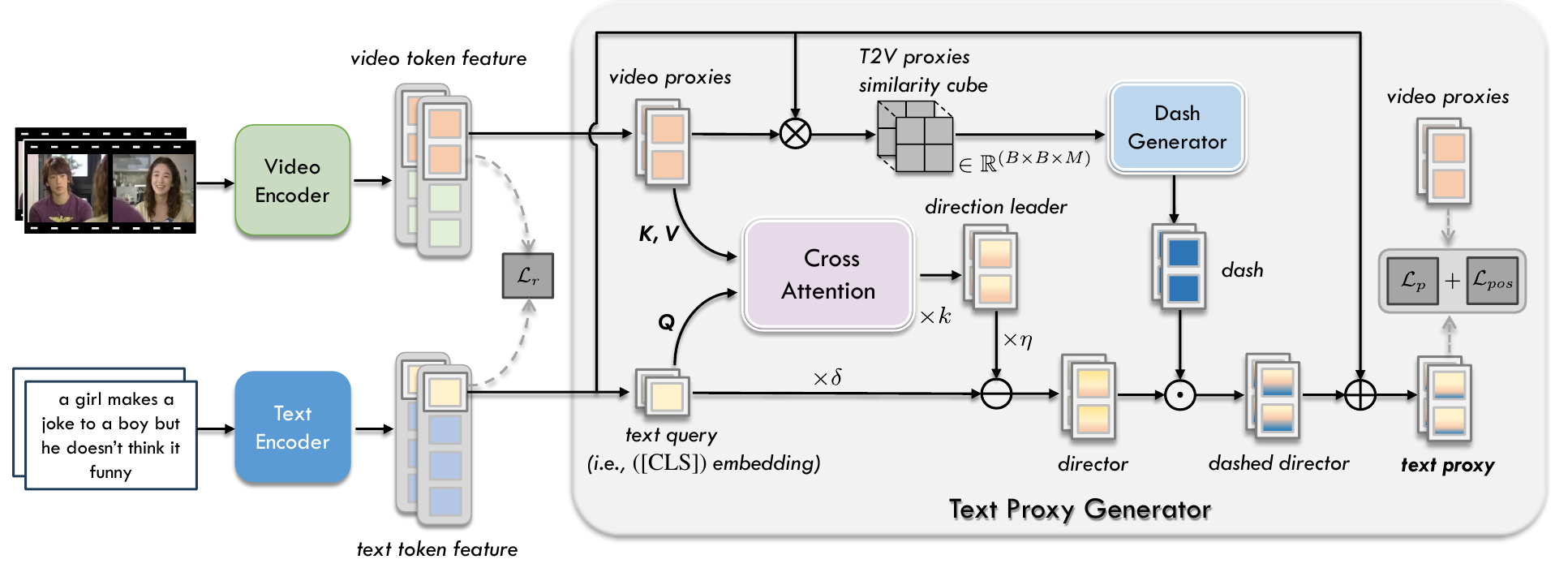}
    \caption{Framework of TV-ProxyNet. After obtaining video proxies and text query from backbone encoders, we  generate direction leader to guide the direction of director and dash to specify the distance between text query and text proxy. During the training stage, $\mathcal{L}_p$ aims to optimize the mentioned $N$ 1-to-1 relationships while $\mathcal{L}_{pos}$ aims to regulate the positive proxy.}
    \label{fig:3}
\end{figure*}

Many CLIP-based methods perform contrastive learning by learning a single holistic video feature through pooling sampled frames to align with text query embeddings. CLIP4Clip \cite{luo2022clip4clip} obtains a video feature by mean pooling all sampled frames, simply transferring a pre-trained image-language model to the video-text domain. To capture higher semantic relevance, X-Pool \cite{gorti2022x} uses cross-attention to generate text-guided video features with varying frame weights. TS2-Net \cite{liu2022ts2} captures fine-grained temporal cues by shifting tokens across frames, achieving strong performance. Due to redundant information in videos, textual data often corresponds to specific segments, leading to correspondence ambiguity. TMVM \cite{lin2022text} and ProST \cite{li2023progressive} address this by adaptively aligning fine-grained prototypes. HBI \cite{jin2023video} proposes hierarchical Banzhaf interaction for interpretable cross-modal comparison of video frames and text words. Unlike these methods focused on multi-grained representation learning, we aim for the model to learn a text-guided proxy for the basic text query embedding, using it to assist in text-video retrieval without multi-grained alignment.

\subsection{Discriminative Text Representation Learning}

Current TVR methods often rely on simplistic video captioning datasets, limiting the discriminative power of text for video retrieval. Cap4Video~\cite{wu2023cap4video} generates captions for data augmentation, enhancing text-query discrimination. \cite{yang2024exploring} improves text features in image captioning using similarity-based image retrieval (SIIR) for better caption accuracy. \cite{li2024configure} leverages CLIP to select similar image triplets, improving visual task recognition. \cite{yang2024lever} enhances text features by configuring ICD sequences, boosting CIDEr scores in image captioning and VQA accuracy. CLIP-ViP \cite{xue2023clip} enriches CLIP by incorporating diverse language sources through large-scale video-text datasets. T-MASS \cite{wang2024text} extends the semantic range of a text query from a point to a mass. Unlike T-MASS, which uses stochastic text embeddings, the proposed method learns a text proxy path and utilizes the final proxy to improve text discriminability by converting the 1-to-$N$ relationship into $N$ 1-to-1 relationships.

\section{Method}

The training objective of TV-ProxyNet is to learn a joint space that minimizes distances between positive pairs and maximizes those between negative pairs. To alleviate the Modeling Irregularity problem, we adopt proxy-to-video retrieval to assist text-to-video retrieval. 
Our goal is for the proxies to have clearer distance distinctions compared to those between the original text queries and videos, thereby increasing the logit score gap between positive and negative pairs.
Effective regulation of proxy distribution requires precise control of each proxy's direction and distance, arranged by the director and dash mechanisms.
The director is a directional vector determined by the direction leader and the query. The illustration of our framework is shown in Fig.~\ref{fig:2}

\subsection{Framework}
TV-ProxyNet consists of a backbone network and a Text Proxy Generator module. Given a raw video, we uniformly sample $T$ frames, resulting in the video input $X=\{x_1,x_2,...,x_T\}$. Each frame includes $H\times W$ patch tokens, where $H$ and $W$ represent the number of patch token in height and width direction respectively. All patch tokens are flattened to obtain a token sequence $X=\{p_i\}_{i=1}^{T\times H\times W}$, where $p_i$ is a patch token. After that, we adopt the same approach as CLIP-ViP~\cite{xue2023clip} to obtain $M$ learnable video proxy tokens and concatenate them to the beginning of $X$. The concatenated token sequence $X_{concat} = \{p_i\}_{i=1}^{M+T\times H\times W}$  is the vision input of backbone network to attend attention score calculation within a video. Also, we have text input $Y=\{t_q,t_1,t_2,...,t_w\}$, where $t_q$ is text ([CLS]) token and $w$ is the number of words. We denote video encoder and text encoder as $\phi$ and $\psi$ respectively. The encoder outputs are:
\begin{equation}
    \mathbf{X}_{concat} = \phi(X_{concat}); \mathbf{Y}=\psi(Y).
\end{equation}
We choose the first video proxy embedding $\mathbf{p}_1 \in \mathbb{R}^{1\times d}$ from the $\mathbf{X}_{concat}$ as the video feature, and choose the text ([CLS]) embedding $\mathbf{t}_q \in \mathbb{R}^{1\times d}$ from $\mathbf{Y}$ as the text feature, where $d$ is the dimension of joint space. After calculating the regular text-video pair contrastive loss, we use $\mathbf{t}_q$ to generate its proxy embedding with the guidance of all $M$ video proxy embeddings. For simplicity, we  refer to the vision encoder's output $\mathbf{X}_{concat}$ as $\mathbf{X}$ in the subsequent illustration.

\subsection{Text Proxy Generation}
The text proxy generation is associated whith the direction and distance of text proxy relative to the text query. 

Given a text query $\mathbf{t}_q$, we generate a direction leader $\mathbf{d}_l$ with arbitrary video in a batch including $N$ text-video pairs, preliminarily establishing $N$ 1-to-1 relationships. Each $\mathbf{d}_l$ specify a direction by the director vector $\mathbf{d}$ with $\mathbf{t}_q$. The dash $\mathcal{D}_s$ is generated by calculating similarities between $\mathbf{t}_q$ and $M$ video proxies $\tilde{\mathbf{X}}=\{\mathbf{p}_1,\mathbf{p}_2,...,\mathbf{p}_M\}$ and projecting them into a scalar, representing the distance between the proxy and the text query. The effect of this process is elaborated in detail in Fig.~\ref{fig:1}(b). The generation of the direction leader and dash forms the core component of TV-ProxyNet. They determine the relative position of proxy to the text query in the joint space. We formulate of generation of proxy $\mathbf{t}_p$ as:

\begin{equation}
\mathbf{t}_p = \mathbf{t}_q + \mathcal{D}_s \cdot \frac{\mathbf{d}}{|\mathbf{d}|},
\end{equation}where $\mathbf{t}_p,\mathbf{d} \in \mathbb{R}^{1\times d}$, $\mathcal{D}_s \in \mathbb{R}^+$.


\subsubsection{Director Generation}
Director $\mathbf{d}$ is the direction vector pointing from text query $\mathbf{t}_q$ to direction leader $\mathbf{d}_l$ or vice versa. To generate director, we need to generate direction leader first. As we iteratively generate $k$ rounds direction leader to form a path, we denote the final direction leader on the path as $\mathbf{d}_{l_k}$. 

Given a text query $\mathbf{t}_q$ and $N$ videos $\tilde{\mathbf{X}}_B=\{\tilde{\mathbf{X}}_j\}_{j=1}^N \in \mathbb{R}^{N\times (M\times d)}$, we duplicate $\mathbf{t}_q$ into $\mathbf{D}_{l_0} = \{\mathbf{t}_q\}_{j=1}^N$ and consider $\mathbf{D}_{l_0}$ as $\mathbf{Q}$, and $\tilde{\mathbf{X}}_B $ as $\mathbf{K},\mathbf{V}$ of cross attention for generating direction leaders $\mathbf{D}_{l_k}=\{\mathbf{d}_{l_k, j}\}_{j=1}^{N} \in \mathbb{R}^{N\times (1\times d)}$. Next we have the $i$-th iteration of direction leader $\mathbf{D}_{l_i}$:
\begin{equation}
\mathbf{D}_{l_i} = \text{softmax}(\mathbf{Q}_{i-1}\cdot \mathbf{K}_{i-1}^\top)\cdot \mathbf{V}_{i-1} + \mathbf{Q}_{i-1},
\end{equation}
where $i\in [1,k]$, $\mathbf{Q}_{i-1}$ and $\mathbf{K}_{i-1},\mathbf{V}_{i-1}$ are the $(i-1)$-th linear transformation of $\mathbf{D}_{l_{i-1}}$ and $\tilde{\mathbf{X}}_B$ respectively, and $\mathbf{K}_{i-1}^\top \in \mathbb{R}^{N\times (d\times M)}$. This operation generates $N$ direction leaders $\mathbf{d}_{l_k}$ for each of text query $\mathbf{t}_q$ in this batch, preliminarily establishing $N$ 1-to-1 relationships.

Generating a direction leader should consider the spatial distribution of $M$ video proxy embeddings containing various semantic information of the video. However, the position of direction leader shouldn't be merely geometric center of $M$ video proxy embeddings but additionally considering semantic of text query. That's why we denote text query as $\mathbf{Q}$ of cross attention to calculate attention score. For $\mathbf{D}_{l_{i-1}}$, it needs to be projected into $\mathbf{Q}_{i-1}$, which means $\mathbf{D}_{l_{i-1}}$ acts as a text guide of the generation of $\mathbf{D}_{l_i}$. Then the last generated direction leader $\mathbf{D}_{l_k}$ takes responsibility to generate the director $\mathbf{d}$ with $\mathbf{t}_q$.

During the training stage, direction leader $\mathbf{d}_{l_k}$ and text query $\mathbf{t}_q$ jointly indicate in which direction the text proxy $\mathbf{t}_p$ should be generated relative to the text query $\mathbf{t}_q$. Thus we have director $\mathbf{d}$:
\begin{equation}
\mathbf{d} = \delta \mathbf{t}_q - \eta \mathbf{d}_{l_k},
\end{equation}
where $\delta$ and $\eta$ are hyperparameters to determine the pointing direction and scale the range of absolute position of $\mathbf{t}_q$ and $\mathbf{d}_{l_k}$. We scale the range to control the dominant of either one. Since the number of patch tokens under ViT-B/16 is much larger than that under ViT-B/32, directly weighting same importance on both of $\mathbf{t}_q$ and $\mathbf{d}_{l_k}$ lacks of specificity under ViT-B/16. Specifically, when $|\delta|$ is lower and $|\eta|$ is higher, the absolute position of $\mathbf{t}_q$ in joint space will be proportionally scaled relative to the origin, resulting in a director $\mathbf{d}$ relatively dominated by $\mathbf{d}_{l_k}$. The lower $|\delta|$, the greater the dominant role of  $\mathbf{d}_{l_k}$, or vice versa. In a word, $\mathbf{d}$ indicates that $\eta \mathbf{d}_{l_k}$ points at $\delta \mathbf{t}_q$ when $\delta > 0$ and $\eta > 0$. We will further discuss the influence of both of these hyperparameters in ablation study.

\subsubsection{Dash Generation}
Dash $\mathcal{D}_s$ is the relative distance between proxy and text query. Since the substantial disparities in data distribution characteristics among diverse datasets, we adopt different dash generation options for different datasets. 

Specifically, given text query $\mathbf{t}_q$ and $M$ video proxy embeddings $\tilde{\mathbf{X}}=\{\mathbf{p}_1,\mathbf{p}_2,...,\mathbf{p}_M\}$, $\mathcal{D}_s$ is generated with the guidance of $M$ similarities between $\mathbf{t}_q$ and $\{\mathbf{p}_1,\mathbf{p}_2,...,\mathbf{p}_M\}$. Denote the similarity vector as $\mathbf{S} = [s_1, s_2,...s_M], s_i = s(\mathbf{t}_q, \mathbf{p}_i)$, where $s(\cdot)$ is cosine similarity function. Then we have dash formulation:
\begin{equation}
    \mathcal{D}_s = \mathrm{exp}(\frac{1}{M}\sum_{i} \theta \cdot s_i),
\end{equation}
where $\theta$ is a learnable scaling factor to scale the mean similarity. And this operation considers all $M$ similarities to be of equal importance. The exponential operation projects the scalar into $\mathbb{R}^+$ as we only need director to specify the direction and dash to specify the distance, meanwhile, it provides higher flexibility. Also, for further enhancing the flexibility, we introduce a dash variant:
\begin{equation}
    \mathbf{d}_s = \mathrm{exp}(\mathbf{SW}),
\end{equation}where $\mathbf{d}_s \in \mathbb{R}_{+}^{1\times d}$ and $\mathbf{W} \in \mathbb{R}^{M\times d}$. $\mathbf{W}$ is a learnable linear projection matrix, aiming to project $M$ similarities into a $d$-dimension vector $\mathbf{d}_s$, which indicates that we allow the proxy to undergo minor disturbances along a specific direction, while each similarity has different weight. 

\subsection{Training Objective}
We adopt the baseline's approach: select the first video proxy embedding $\mathbf{p}_1$ to represent the video $\mathbf{X}$. During the training stage, we expect the distance between the $i$-th video $\mathbf{p}_{1_i}$ and the text proxy $\mathbf{t}_{p_{i,i}}$ generated by the $i$-th text query and the $i$-th video should be closer than that between the $j$-th video $\mathbf{p}_{1_j}$ and the proxy $\mathbf{t}_{p_{i,j}}$, which is similar to the regular text-video pair training objective. The distinction lies in our utilization of proxy-video pair as a substitute for text-video pair to execute contrastive learning. Next we have the proxy-video pair infoNCE loss \cite{oord2018representation}:
\begin{equation}
    \mathcal{L}_{p}^{p2v} = -\frac{1}{B}\sum_{i=1}^{B}\mathrm{log}\frac{\mathrm{exp}(s(\mathbf{t}_{p_{i,i}}, \mathbf{p}_{1_i})/\sigma)}{\sum_{j=1}^{B} \mathrm{exp}(s(\mathbf{t}_{p_{i,j}}, \mathbf{p}_{1_j})/\sigma)},
\end{equation}
\begin{equation}
   \mathcal{L}_{p}^{v2p} = -\frac{1}{B}\sum_{i=1}^{B}\mathrm{log}\frac{\mathrm{exp}(s(\mathbf{p}_{1_i}, \mathbf{t}_{p_{i,i}})/\sigma)}{\sum_{j=1}^{B} \mathrm{exp}(s(\mathbf{p}_{1_j}, \mathbf{t}_{p_{i,j}})/\sigma)},
\end{equation}
\begin{equation}
    \mathcal{L}_p = \frac{1}{2}(\mathcal{L}_{p}^{p2v} + \mathcal{L}_{p}^{v2p}),
\end{equation}where $\sigma$ is learnable temperature and $B$ is batch size.

For the positive text proxy $\mathbf{t}_{p_{i,i}}$, not only the distance between $\mathbf{t}_{p_{i,i}}$ and $\mathbf{p}_{1_i}$ should be closer than that between the negative proxy $\mathbf{t}_{p_{i,j}}$ and $\mathbf{p}_{1_j}$, but also the distance between $\mathbf{t}_{p_{i,i}}$ and $\mathbf{p}_{1_i}$ should be closer than that between $\mathbf{t}_{p_{i,i}}$ and $\mathbf{p}_{1_j}$. Therefore, we need to establish a hard negative constraint for this type of proxy to further regulate the direction and distance of positive proxy. In this regard, we have positive proxy infoNCE loss:

\begin{equation}
    \mathcal{L}_{pos}^{p2v} = -\frac{1}{B}\sum_{i=1}^{B}\mathrm{log}\frac{\mathrm{exp}(s(\mathbf{t}_{p_{i,i}}, \mathbf{p}_{1_i})/\sigma)}{\sum_{j=1}^{B} \mathrm{exp}(s(\mathbf{t}_{p_{i,i}}, \mathbf{p}_{1_j})/\sigma)},
\end{equation}
\begin{equation}
   \mathcal{L}_{pos} = \frac{1}{2} (\mathcal{L}_{pos}^{p2v} + \mathcal{L}_{pos}^{v2p}).
\end{equation}Since we require a text query to locate the proxy, we first ensure that the text query is correctly distributed. If the text query is not properly distributed, the proxy path cannot be normalized. Therefore, we employ the infoNCE loss of regular text-video pairs:
\begin{equation}
    \mathcal{L}_{r}^{t2v} = -\frac{1}{B}\sum_{i=1}^{B}\mathrm{log}\frac{\mathrm{exp}(s(\mathbf{t}_{q_i}, \mathbf{p}_{1_i})/\sigma)}{\sum_{j=1}^{B} \mathrm{exp}(s(\mathbf{t}_{q_j}, \mathbf{p}_{1_j})/\sigma)},
\end{equation}
\begin{equation}
    \mathcal{L}_r = \frac{1}{2}(\mathcal{L}_{r}^{t2v} + \mathcal{L}_{r}^{v2t}).
\end{equation}Finally, the total loss is 
\begin{equation}
    \mathcal{L} = \mathcal{L}_{r} + \alpha \mathcal{L}_p + \beta \mathcal{L}_{pos}.
\end{equation}
where $\alpha$ and $\beta$ are hyperparameters that weigh the distribution characteristics of the text query and text proxy within the joint space.

\begin{table*}[!tbp]
    \centering
    \begin{tabularx}{\textwidth}{l|>{\centering\arraybackslash}X>{\centering\arraybackslash}X>{\centering\arraybackslash}X>{\centering\arraybackslash}X>{\centering\arraybackslash}X|>{\centering\arraybackslash}X>{\centering\arraybackslash}X>{\centering\arraybackslash}X>{\centering\arraybackslash}X>{\centering\arraybackslash}X} 
    \Xhline{1pt}
    \multirow{2}*{Method} & \multicolumn{5}{c|}{MSRVTT Retrieval} & \multicolumn{5}{c}{DiDeMo Retrieval} \\
    \Xcline{2-11}{0.4pt}
    & R@1 & R@5 & R@10 & MdR$\downarrow$ & MnR$\downarrow$ & R@1 & R@5 & R@10 & MdR$\downarrow$ & MnR$\downarrow$ \\
    \Xhline{0.4pt}
    {\small \textit{CLIP-ViT-B/32}} & \multicolumn{5}{c|}{} & \multicolumn{5}{c}{}\\ 
    EMCL~\cite{jin2022expectation} & 46.8 & 73.1 & 83.1 & 2.0 & - & - & - & - & - & - \\
    X-Pool~\cite{gorti2022x} & 46.9 & 72.8 & 82.2 & 2.0 & 14.3 & 44.6 & 73.2 & 82.0 & 2.0 & 15.4 \\
    TS2-Net~\cite{liu2022ts2} & 47.0 & 74.2 & 83.3 & 2.0 & 13.6 & 41.8 & 71.6 & 82.0 & 2.0 & 14.8 \\
    UATVR~\cite{fang2023uatvr} & 47.5 & 73.9 & 83.5 & 2.0 & 12.3 & 43.1 & 71.8 & 82.3 & 2.0 & 15.1 \\
    ProST~\cite{li2023progressive} & 48.2 & 74.6 & 83.4 & 2.0 & 12.4 & 44.9 & 72.7 & 82.7 & 2.0 & 13.7 \\
    HBI~\cite{jin2023video} & 48.6 & 74.6 & 83.4 & 2.0 & 12.0 & 46.9 & 74.9 & 82.7 & 2.0 & 12.1 \\
    Cap4Video\textdaggerdbl~\cite{wu2023cap4video} & 49.3 & 74.3 & 83.8 & 2.0 & 12.0 & \textbf{52.0} & \textbf{79.4} & \textbf{87.5} & 1.0 & \textbf{10.5} \\
    CLIP-ViP\textdaggerdbl~\cite{xue2023clip} & 50.1 & 74.8 & 84.6 & 1.0 & - & 48.6 & 77.1 & 84.4 & 2.0 & - \\
    T-MASS~\cite{wang2024text} & \underline{50.2} & \underline{75.3} & \underline{85.1} & 1.0 & \underline{11.9} & \underline{50.9} & \underline{77.2} & 85.3 & 1.0 & 12.1 \\
    TV-ProxyNet~(Ours) & \textbf{52.3} & \textbf{77.8} & \textbf{85.8} & 1.0 & \textbf{11.1} & 50.6 & 76.9 & \underline{86.0} & 1.0 & \underline{11.5} \\
    \Xhline{0.4pt}
    {\small \textit{CLIP-ViT-B/16}} & \multicolumn{5}{c|}{} & \multicolumn{5}{c}{}\\
    X-Pool~\cite{gorti2022x} & 48.2 & 73.7 & 82.6 & 2.0 & 12.7 & 47.3 & 74.8 & 82.8 & 2.0 & 14.2 \\
    ProST~\cite{li2023progressive} & 49.5 & 75.0 & 84.0 & 2.0 & 11.7 & 47.5 & 75.5 & 84.4 & 2.0 & 12.3 \\
    UATVR~\cite{fang2023uatvr} & 50.8 & 76.3 & 85.5 & 1.0 & 12.4 & 45.8 & 73.7 & 83.3 & 2.0 & 13.5 \\
    CLIP-ViP\textdaggerdbl~\cite{xue2023clip} & \underline{54.2} & \underline{77.2} & 84.8 & 1.0 & - & 50.5 & 78.4 & 87.1 & 1.0 & - \\ 
    Cap4Video\textdaggerdbl~\cite{wu2023cap4video} & 51.4 & 75.7 & 83.9 & 1.0 & 12.4 & - & - & - & - & - \\
    T-MASS~\cite{wang2024text} & 52.7 & 77.1 & \underline{85.6} & 1.0 & \underline{10.5} & \textbf{53.3} & \underline{80.1} & \underline{87.7} & 1.0 & \underline{9.8} \\
    TV-ProxyNet~(Ours) & \textbf{55.2} & \textbf{80.4} & \textbf{86.8} & 1.0 & \textbf{9.3} & \underline{52.1} & \textbf{80.4} & \textbf{88.7} & 1.0 & \textbf{8.7} \\
    \Xhline{1pt}
    \end{tabularx}
    \caption{Text-to-video Retrieval comparisons on MSRVTT and DiDeMo. Bold denotes the best performance. Underline denotes the second performance. “-” result is unavailable. \textdaggerdbl ~denotes the method using extra data.}
    \label{tab:table1}
\end{table*}

\subsection{Inference Pipeline}
During the inference stage, we employ proxy-video logit score $s(\mathbf{t}_{p_{i,j}}, \mathbf{p}_{1_j})$ to remedy text-video logit score $s(\mathbf{t}_{q_i}, \mathbf{p}_{1_j})$. This alters the logit scores for both positive and negative text-video sample pairs, aiming to enhance the likelihood of retrieving the ground-truth videos. Thus, the final retrieval logit score is $s(\mathbf{t}_{q_i}, \mathbf{p}_{1_j}) + \gamma s(\mathbf{t}_{p_{i,j}}, \mathbf{p}_{1_j})
$, where $\gamma \in [0.1, 0.8]$ is a weight coefficient. Notice that this logit score is no longer a simple cosine similarity but rather a linear combination of different cosine similarities. An ablation study of $\gamma$, a more detailed analysis of the formula derivation and visualization are provided in the appendix.

It's crucial to note that we avoid computing the logit score $s(\mathbf{t}_{p_{i,i}}, \mathbf{p}_{1_j})$ for remedying the logit score of $(i,j)$-th text-video pair, as this would constitute cheating. This is because the positive proxy $\mathbf{t}_{p_{i,i}}$ is derived from the $i$-th video, inevitably leading to a higher similarity.

\section{Experiment}
\subsection{Datasets and Implementation}
We employ three TVR benchmarks for evaluation:
(1) \textbf{MSRVTT} \cite{xu2016msr} consists of 10k videos and each video has 20 captions. We train 9k train+val videos and test on a test set of 1k text-video pairs. (2) \textbf{DiDeMo} \cite{ anne2017localizing} contains 10K Flickr videos and 40K captions. All caption descriptions of a video are concatenated as a query to perform paragraph-video retrieval. (3) \textbf{ActivityNet Captions} \cite{krishna2017dense} contains 20K YouTube videos annotated with 100K sentences. We follow the paragraph-video retrieval setting to train models on 10K videos and report results on the val1 set with 4.9K videos.

We employ CLIP-ViP~\cite{xue2023clip} (both ViT-B/32 and ViT-B/16) as our backbone.  All videos are resized to 224$\times$224. We set frame number to 12 for a fair comparison except for ActivityNet (set to 32) as its videos are much longer. The text proxy generator module consists of $k$ Transformer layers with a single attention head and a linear dash generator. We finally set $k$ to 2 to perform two rounds direction leader generation. For learning rate, we set backbone to 1e-6 and text proxy generator module to 2e-6 for MSRVTT, 2e-6 and 4e-6 for DiDeMo and ActivityNet. We optimize the model for 5 epochs for MSRVTT on 2 RTX 4090 GPUs, 20 epochs for DiDeMo on 1 RTX 4090 GPU and ActivityNet on an A800 GPU, employ AdamW optimizer with weight decay 0.2. The training batchsize is 128 for MSRVTT, 64 for DiDeMo and ActivityNet. The $\alpha$ and $\beta$ hyperparameters are (0.5, 0.25) for MSRVTT, (1, 0.25) for DiDeMo and  (0.75, 0.25) for ActivityNet Captions. The dimension of proxy is 512 same as the backbone's output.

\subsection{Comparison with State-of-the-art Methods}

\begin{table}[t]
    \centering
    \begin{tabularx}{\linewidth}{l|>{\centering\arraybackslash}X>{\centering\arraybackslash}X>{\centering\arraybackslash}X>{\centering\arraybackslash}X}  
    \Xhline{1pt}
    Method & \small R@1 & \small R@5 & \small R@10 & \small MdR \\
    \hline 
    {\small \textit{CLIP-ViT-B/32}} & \multicolumn{4}{c}{}\\
    CLIP4Clip~\cite{luo2022clip4clip} & 40.5 & 72.4 & - & 2.0 \\
    EMCL~\cite{jin2022expectation} & 41.2 & 72.7 & - & 2.0 \\ 
    HBI~\cite{jin2023video} & 42.2 & 73.0 & 84.6 & 2.0 \\
    CenterCLIP~\cite{zhao2022centerclip} & 43.9 & 74.6 & 85.8 & 2.0 \\
    CLIP2TV~\cite{gao2021clip2tv}& 44.1 & 75.2 & - & 2.0 \\
    DRL~\cite{DRLTVR2022} & 44.2 & 74.5 & 86.1 & 2.0 \\
    CAMoE*~\cite{cheng2021improving} & 51.0 & 77.7 & - & - \\
    CLIP-ViP\textdaggerdbl~\cite{xue2023clip} & \underline{51.1} & \underline{78.4} & \underline{88.3} & 1.0 \\
    TV-ProxyNet~(Ours) & \textbf{53.0} & \textbf{80.9} & \textbf{89.6} & 1.0 \\
    \hline
    {\small \textit{CLIP-ViT-B/16}} & \multicolumn{4}{c}{}\\
    CenterCLIP~\cite{zhao2022centerclip} & 46.2 & 77.0 & 87.6 & 2.0 \\
    DRL~\cite{DRLTVR2022} & 46.2 & 77.3 & 88.2 & 2.0 \\
    CLIP-ViP\textdaggerdbl~\cite{xue2023clip} & \underline{53.4} & \underline{81.4} & \underline{90.0} & 1.0 \\
    TV-ProxyNet~(Ours) & \textbf{56.7} & \textbf{83.1} & \textbf{91.1} & 1.0 \\
    \Xhline{1pt}
    \end{tabularx}
    \caption{Text-to-video Retrieval comparisons on ActivityNet Captions. Bold denotes the best performance. Underline denotes the second performance. “-” result is unavailable. * denotes that the use of DSL as post-processing operation. \textdaggerdbl ~denotes the method using extra data.}
    \label{tab:table2}
\end{table}

We compare the performance of TV-ProxyNet with recent state-of-the-art methods and list the results in Tab.~\ref{tab:table1} and Tab.~\ref{tab:table2}. We have achieved the state-of-the-art performance on certain metrics of two datasets. On MSRVTT, TV-ProxyNet based on ViT-B/32 improves the R@1 metric by 2.2\% (52.3\% v.s. 50.1\%) compared to CLIP-ViP, while based on ViT-B/16, it improves the R@5 metric by 3.2\% (80.4\% v.s. 77.2\%). Meanwhile, under DiDeMo, although we did not exceed T-MASS, we gain the improvement of 2.0\% (50.6\% v.s. 48.6\%) under ViT-B/32 compared to CLIP-ViP. As shown in Tab.~\ref{tab:table2}, under ActivityNet Captions, TV-ProxyNet has improved by 1.9\% (53.0\% v.s. 51.1\%) under ViT-B/32 on R@1 metric compared to the baseline while under ViT-B/16 it has improved by 3.3\% (56.7\% v.s. 53.4\%). It can be seen that the proposed method has a certain degree of continuous performance improvement on different datasets. It should be explained that due to the limitation of experimental resources, we cannot use the learning rate (1e-6) provided in the baseline configuration file to reproduce the DiDeMo results provided in the baseline article. This may be due to the large batch size of a single GPU and the small learning rate. Therefore, we changed the learning rate of the backbone to 2e-6 (batch size 64 on ViT-B/32) on DiDeMo to obtain results close to the baseline. The reproduction result on DiDeMo is R@1=48.3, which indicates that we can achieve a performance improvement of 2.3\% (50.6\% vs.48.3\%) on this dataset using text proxy.

Compared to T-MASS, which randomly generated text embedding with a mass range, TV-ProxyNet improves the R@1 metrics on ViT-B/32 and ViT-B/16 by 2.1\% and 2.0\% on MSRVTT respectively. It is worthy noting that, T-MASS still establishes a 1-to-$N$ relationship for a given text query in the training stage. Instead, TV-ProxyNet aims to exaggerate the relative distances between a given text query and all videos through the distribution of the proxies, thereby improving the likelihood of retrieving the target video, which is implemented by establishing $N$ 1-to-1 relationships. 

\subsection{Ablation Studies}
We conducted the  ablation experiments on different variables of the Texy Proxy Generator module on MSRVTT-1k. The first two ablation experiments are conducted under ViT-B/32, and the last is under ViT-B/16.

\begin{table}[!tbp]
    \centering
    \begin{tabularx}{\linewidth}{ccc|>{\centering\arraybackslash}X>{\centering\arraybackslash}X>{\centering\arraybackslash}X>{\centering\arraybackslash}X}  
    \Xhline{1pt}
    $\mathcal{L}_r$ & $\mathcal{L}_p$ & $\mathcal{L}_{pos}$ & R@1 & R@5 & R@10 & MnR \\
    \hline 
    \ding{51} & \multicolumn{2}{c|}{} & 50.1 & 74.8 & 84.6 & - \\
    \ding{51} & \ding{51} & \multicolumn{1}{c|}{} & 52.0 & 77.7 & 85.6 & \textbf{11.1} \\
    \ding{51} &  & \multicolumn{1}{c|}{\ding{51}} & 50.6 & 76.5 & 85.2 & 11.3 \\
    \ding{51} & \ding{51} & \multicolumn{1}{c|}{\ding{51}} & \textbf{52.3} & \textbf{77.8} & \textbf{85.8} & \textbf{11.1} \\
    \Xhline{1pt} 
    \end{tabularx}
    \caption{The ablation study of different losses on MSRVTT-9k. Row 1 only using $\mathcal{L}_r$ refers to baseline.}
    \label{tab:table_losses}
\end{table}

\subsubsection{The Effect of Losses}
We evaluated the impact of adding two proposed training objectives to the baseline, as shown in Tab.~\ref{tab:table_losses}. The baseline (Row 1) uses only $\mathcal{L}_r$. Adding $\mathcal{L}_p$ (Row 2) significantly improves performance by aligning positive proxy-video pairs more closely while separating negative pairs. This highlights how $\mathcal{L}_p$ effectively decomposes the 1-to-$N$ relationship into $N$ 1-to-1 relationships, reducing interference and increasing flexibility in managing relative distribution.

The best results (Row 4) are achieved by combining $\mathcal{L}_p$ and $\mathcal{L}_{pos}$, confirming that these objectives together enhance the $N$ 1-to-1 framework’s effectiveness. In contrast, adding only $\mathcal{L}_{pos}$ (Row 3) yields minimal improvement, as it regulates positive proxies without transitioning to the 1-to-$N$ decomposition. Notably, in Row 3, we still use proxy-video logits to complement text-video logits during inference, assigning varying proportions by $\gamma$. However, performance peaks when $\gamma = 0$, suggesting that $\mathcal{L}_{pos}$ works best when paired with $\mathcal{L}_p$.

\subsubsection{The Effect of Dash and Iteration of Direction Leader}
To evaluate the impact of the dash mechanism, we conducted ablation experiments under three settings: $\mathrm{exp}(\frac{1}{M}\sum \theta \cdot s_i)$, $\mathrm{exp}(\mathbf{SW})$, and without dash (i.e., using the direction leader as the text proxy). 

As shown in Tab.~\ref{tab:table_D}, generating proxies consistently outperformed the baseline, especially under $\mathrm{exp}(\frac{1}{M}\sum \theta \cdot s_i)$. Optimal performance was observed after three iterations with this setting, where improvements stabilized as iterations increased. However, $\mathrm{exp}(\mathbf{SW})$ showed no consistent pattern across iterations. This suggests that while the dash generator has fewer parameters than a single Transformer layer, its design significantly affects proxy distribution. Simpler dash generation methods, such as $\mathrm{exp}(\frac{1}{M}\sum \theta \cdot s_i)$, introduce fewer uncontrollable factors, improving both the controllability of direction leader generation and proxy distribution.

\begin{table}[!tbp]
    \centering
    \begin{tabularx}{\linewidth}{c|>{\centering\arraybackslash}X>{\centering\arraybackslash}X>{\centering\arraybackslash}X>{\centering\arraybackslash}X}  
    \Xhline{1pt}
    \multirow{2}*{Round(s)} & \multicolumn{4}{c}{\small$\mathrm{exp}(\sum \theta \cdot s_i/M)$}  \\
    \Xcline{2-5}{0.4pt}
    & \small R@1 & \small R@5 & \small R@10 & \small MnR  \\
    \hline 
    \multicolumn{1}{c|}{1} & 51.9 & 77.5 & 85.3 & 11.5  \\
    \multicolumn{1}{c|}{2} & 52.3 & 77.8 & \textbf{85.8} & 11.1 \\
    \multicolumn{1}{c|}{3} & \textbf{52.5} & 77.8 & 85.4 & 11.1 \\
    \multicolumn{1}{c|}{4} & 52.4 & 77.6 & \textbf{85.8} & 11.0 \\
    \multicolumn{1}{c|}{5} & 52.3 & 77.4 & 85.6 & 11.1 \\
    \hline 
    \multirow{2}*{Round(s)} & \multicolumn{4}{c}{\small$\mathrm{exp}(\mathbf{SW})$}  \\
    \Xcline{2-5}{0.4pt}
    &\small  R@1 &\small R@5 &\small R@10 &\small MnR  \\
    \hline 
    \multicolumn{1}{c|}{1} & 52.2 & \textbf{77.9} & 85.3 & 11.2 \\
    \multicolumn{1}{c|}{2} & 51.9 & 77.5 & 85.3 & 11.2 \\
    \multicolumn{1}{c|}{3} & 52.2 & 77.8 & 85.1 & 11.3 \\
    \multicolumn{1}{c|}{4} & 52.0 & 77.6 & 85.2 & 11.0 \\
    \hline
    \multirow{2}*{Round(s)} & \multicolumn{4}{c}{w/o dash}  \\
    \Xcline{2-5}{0.4pt}
    &\small  R@1 &\small R@5 &\small R@10 &\small MnR  \\
    \hline
    \multicolumn{1}{c|}{1} & 51.1 & 76.5 & 84.6 & 11.3 \\
    \multicolumn{1}{c|}{2} & 51.9 & 76.4 & 84.5 & \textbf{10.8} \\
    \multicolumn{1}{c|}{3} & 50.3 & 76.3 & 84.5 & 11.8 \\
    \Xhline{1pt} 
    \end{tabularx}
    \caption{Evaluation of direction leader iteration on MSRVTT-9k. “w/o dash” means we directly adopt direction leader $\mathbf{d}_l$ as text proxy without specifying the distance between proxy and query.}
    \label{tab:table_D}
\end{table}

Since each iteration adds computational cost, we opted for two iterations for efficiency, as this configuration also achieved optimal performance under $\mathrm{exp}(\mathbf{SW})$ on two other datasets. To illustrate the effect of iteration count, we focus on $\mathrm{exp}(\frac{1}{M}\sum \theta \cdot s_i)$. A single iteration exaggerates the text query's relative distance to videos, boosting retrieval accuracy but overfitting the text query's semantics, reducing generalization. Conversely, five iterations overly incorporate video information, weakening the text query's semantic guidance. Three iterations strike a balance, ensuring both guidance and generalization.

To assess the role of dash, we also evaluated proxies using only the direction leader across iteration schemes. As shown in Tab.~\ref{tab:table_D}, this setup performed well after two iterations but lacked scalability with more iterations. Solely relying on the direction leader as the proxy imposes a burden on its generation, as it cannot effectively determine the proxy’s relative distance to the query.

Finally, we adopt $\mathrm{exp}(\frac{1}{M}\sum \theta \cdot s_i)$ for MSRVTT and $\mathrm{exp}(\mathbf{SW})$ for DiDeMo and ActivityNet Captions. We attribute this difference to the concise text information in MSRVTT, which aligns better with the former, while the latter provides more flexibility for the diverse semantics in the other datasets.

\begin{table}[!tbp]
    \centering
    \begin{tabularx}{\linewidth}{c|>{\centering\arraybackslash}X>{\centering\arraybackslash}X>{\centering\arraybackslash}X>{\centering\arraybackslash}X}  
    \Xhline{1pt}
    $\{\delta, \eta\}$  & R@1 & R@5 & R@10 & MnR \\
    \hline 
    \multicolumn{1}{c|}{\{1.5,1\}} & 54.3 & 79.4 & 87.0 & 8.9 \\
    \multicolumn{1}{c|}{\{1,1\}} & 54.2 & 78.9 & \textbf{87.2} & 8.8 \\
    \multicolumn{1}{c|}{\{0.5,1\}} & 54.1 & 79.0 & 86.9 & 9.2 \\
    \hline
    \multicolumn{1}{c|}{\{-1.5,-1\}} & 55.1 & 80.2 & 86.9 & 9.2 \\
    \multicolumn{1}{c|}{\{-1,-1\}} & \textbf{55.2} & \textbf{80.4} & 86.8 & 9.3 \\
    \multicolumn{1}{c|}{\{-0.5,-1\}} & 54.9 & 79.9 & 86.7 & \textbf{8.7} \\
    \Xhline{1pt} 
    \end{tabularx}
    \caption{Evalutation to investigate the effect of $\delta$ and $\eta$ under ViT-B/16 on MSRVTT-9k . Row 1-3 imply the direction leader points at the text proxy, Row 4-6 imply the opposite.}
    \label{tab:table_hypers}
\end{table}

\subsubsection{The Effect of $\delta$ and $\eta$}

We control the dominance and pointing direction between the text query and the direction leader by adjusting $\delta$ and $\eta$, respectively. While performance fluctuations under ViT-B/32 are minor (within 0.3\%), both $\delta$ and $\eta$ significantly influence performance under ViT-B/16. Thus, this ablation experiment focuses on ViT-B/16. Due to resource constraints, our reproduced result on MSRVTT achieves R@1=53.5 (baseline: R@1=54.2) using a batch size of 56 on an A800 GPU.

As shown in Tab.~\ref{tab:table_hypers}, when $\delta > 0$ and $\eta > 0$, the R@1 metric shows limited improvement, as the direction leader primarily points towards the text query. Conversely, with $\delta < 0$ and $\eta < 0$, R@1 increases more significantly. This improvement is likely because ViT-B/16 has a larger number of patch tokens than ViT-B/32, meaning proxies under $\delta > 0$ and $\eta > 0$ are insufficiently exaggerated to enhance text-video logits. For ViT-B/32, we set $\delta = 1$ and $\eta = 1$.

\section{Conclusion}
This work introduces TV-ProxyNet, a novel framework for addressing challenges in text-video retrieval (TVR). By reformulating the task from a traditional 1-to-$N$ matching problem into $N$ 1-to-1 relationships, our method effectively mitigates modeling irregularities. Leveraging text proxies guided by the director and dash mechanisms, TV-ProxyNet achieves precise alignment, enhancing retrieval accuracy and robustness. Extensive evaluations on multiple benchmarks demonstrate its superiority over state-of-the-art methods, offering a robust solution for complex multimodal datasets and paving the way for improved semantic understanding and retrieval in real-world applications.

\section{Acknowledgments}
This work was supported by the NSFC NO. 62172138 and No. 62202139. This work was also partially supported by the Fundamental Research Funds for the Central Universities NO. JZ2024HGTG0310

\bibliography{references}

\input{appendix}

\end{document}

%% file: appendix.tex
\setcounter{secnumdepth}{0} 
\maketitle

\appendix 
\section*{Appendix} 
\addcontentsline{toc}{section}{Appendix}
This appendix provides more details of TV-ProxyNet framework: 1) ablation study on $\gamma$; 2) derivation and analysis of retrieval formula and feature space visualization; 3) complexity and computational cost; 4) retrieval cases analysis.

\subsection{Ablation study on $\gamma$}
\begin{figure}[t]
    \centering
    \includegraphics[width=0.9\linewidth]{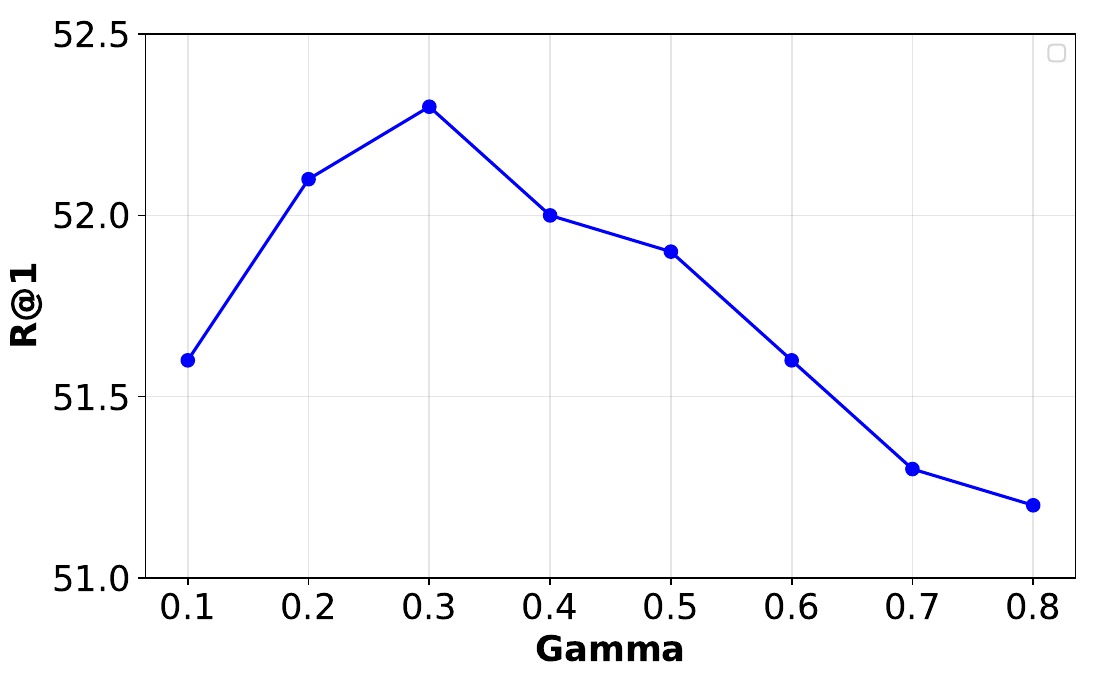}
    \caption{The R@1 results at different $\gamma$.}
    \label{fig:4}
\end{figure}

As shown in Fig.~\ref{fig:4}, we examine the impact of different values of $\gamma$ on the Recall@1 metric at the training step where the best results are achieved. In fact, when considering the inference results across various training steps throughout the entire training process, it becomes apparent that the effect of $\gamma$ on the outcomes is irregular. We hypothesize that this irregularity arises from the varying degrees to which the different learned representational spaces during training adapt to $\gamma$, leading to this unpredictable phenomenon.

\subsection{Analysis and Visualization}

\subsubsection{Analysis of Relationship Decomposing}
Given a query and \( N \) videos, we treat the relationship between the query and each video as distinct. Thus, we decompose the retrieval process from a 1-to-\( N \) relationship into \( N \) 1-to-1 relationships, allowing the model to account for the specificity of each. Specifically, for query $\mathbf{t}_{q_i}$, we generate proxies \( \mathbf{t}_{p_{i,i}} \) and \( \mathbf{t}_{p_{i,j}} \) for videos $\mathbf{p}_{1_i}$ and $\mathbf{p}_{1_j}$, respectively. During loss optimization, the numerator \( \exp(s(\mathbf{t}_{p_{i,i}}, p_{1_i})) \) in the softmax function minimally impacts the denominator \( \exp(s(\mathbf{t}_{p_{i,j}}, p_{1_j})) \), ensuring independent optimization of their parameter sets. This differs from optimizing the numerator \( \exp(s(\mathbf{t}_{q_i}, p_{1_i})) \) of \( \mathcal{L}_r \), which significantly impacts its denominator \( \exp(s(\mathbf{t}_{q_i}, p_{1_j})) \).

\subsubsection{Derivation of Retrieval formula before Visualizing}
Given that the logit scores in our retrieval process are not solely derived from the cosine similarity between features but instead represent a linear combination of two cosine similarities, visualizing the results using only the text proxy lacks sufficient interpretability. Since the proxy functions as an auxiliary component to facilitate query-based retrieval, we extend the retrieval formula to include a linear term (monomial) based on cosine similarity. Specifically, let \( Y_i \) denote the $i$-th raw text input, \( X_j \) the $j$-th raw video input, and \( \text{Sim}(\cdot) \) the similarity function learned by the model, where the similarity between  \( Y_i \) and \( X_j \) is represented as \( \text{Sim}(X_j, Y_i) \), and $\text{Sim}(X_j, Y_i)$ can be extended to 
\begin{align*}
\text{Sim}(X_j, Y_i)
&= s(\mathbf{t}_{q_i}, \mathbf{p}_{1_j}) + \gamma s(\mathbf{t}_{p_{i,j}}, \mathbf{p}_{1_j}) \\
&= \frac{\langle \mathbf{t}_{q_i}, \mathbf{p}_{1_j} \rangle}{|\mathbf{t}_{q_i}|\cdot |\mathbf{p}_{1_j}|} + \gamma \frac{\langle \mathbf{t}_{p_{i,j}}, \mathbf{p}_{1_j} \rangle}{|\mathbf{t}_{p_{i,j}}|\cdot |\mathbf{p}_{1_j}|} \\
&= \frac{\sum_{k=1}^{d} t_{q_i, k} \cdot p_{1_j, k} }{|\mathbf{t}_{q_i}|\cdot |\mathbf{p}_{1_j}|} + \gamma \frac{\sum_{k=1}^{d} t_{p_{i,j}, k} \cdot p_{1_j, k} } {|\mathbf{t}_{p_{i,j}}|\cdot |\mathbf{p}_{1_j}|} \\
&= \sum_{k=1}^d \underbrace{(\frac{t_{q_i, k}}{|\mathbf{t}_{q_i}|} + \gamma \frac{t_{p_{i,j}, k}}{|\mathbf{t}_{p_{i,j}}|})}_{q_{(i,j), k}}\cdot \frac{p_{1_j, k}}{|\mathbf{p}_{1_j}|}   \\
&= \sqrt{\sum_{k=1}^d q_{(i,j), k}^2 }\cdot \sum_{k=1}^d \frac{q_{(i,j), k}}{\sqrt{\sum_{k=1}^d q_{(i,j), k}^2 }} \cdot \frac{p_{1_j, k}}{|\mathbf{p}_{1_j}|} \tag{I}.
\end{align*}

Here, let \(\mathbf{q}_{i,j} = (q_{(i,j),k})_{k=1}^d = (\frac{\mathbf{t}_{q_i}}{|\mathbf{t}_{q_i}|} + \gamma \frac{\mathbf{t}_{p_{i,j}}}{|\mathbf{t}_{p_{i,j}}|})_{k=1}^d\). Thus, we have the self-inner product of $\mathbf{q}_{i,j}$:
\begin{align*}
\sum_{k=1}^d q_{(i,j), k}^2 &= \sum_{k=1}^d \frac{t_{q_i,k}^2}{|\mathbf{t}_{q_i}|^2} + \gamma^2 \sum_{k=1}^d \frac{t_{p_{i,j},k}^2}{|\mathbf{t}_{p_{i,j}}|^2} \\ &\quad \ + 2\gamma \sum_{k=1}^d\frac{t_{q_{i},k}}{|\mathbf{t}_{q_i}|}\frac{t_{p_{i,j},k}}{|\mathbf{t}_{p_{i,j}}|} \\
&= 1 + \gamma^2 + 2\gamma s(\mathbf{t}_{q_i}, \mathbf{t}_{p_{i,j}})
\end{align*}

The result is obtained by continuing from $(\text{I})$, and we finally have 

\begin{align*}
\text{Sim}(X_j,Y_i) = \sqrt{1 + \gamma^2 + 2\gamma s(\mathbf{t}_{q_i}, \mathbf{t}_{p_{i,j}}}) \cdot s(\mathbf{q}_{i,j}, \mathbf{p}_{1_j}),
\end{align*}
where $\mathbf{q}_{i,j} = \frac{\mathbf{t}_{q_i}}{|\mathbf{t}_{q_i}|} + \gamma \frac{\mathbf{t}_{p_{i,j}}}{|\mathbf{t}_{p_{i,j}}|}$ is computed externally to the model. This representation $\mathbf{q}_{i,j}$, together with the video $\mathbf{p}_{1,j}$ and query $\mathbf{t}_{q_i}$, is then utilized to visualize the feature space learned by the model.

Note that the square root term's coefficient is determined by $\gamma$ and the similarity \( s(\mathbf{t}_{q_i}, \mathbf{t}_{p_{i,j}}) \), and it is always positive. Therefore, for $(i,i)$-th pair and $(i,j)$-th pair, this coefficient scales the newly obtained similarity \( s(\mathbf{q}_{i,i}, \mathbf{p}_{1_i}) \) and $s(\mathbf{q}_{i,j}, \mathbf{p}_{1_j})$ within the positive range. Below, we focus exclusively on cases where retrieval errors occur in the baseline, i.e., for any $i,j$ such that \(s(\mathbf{t}_{q_i}, \mathbf{p}_{1_i}) \leq s(\mathbf{t}_{q_i}, \mathbf{p}_{1_j})\). If \(\text{Sim}(Y_i, X_i) \geq \text{Sim}( Y_i, X_j)\), it's obviously that the learned \(\mathbf{t}_{p_{i,i}}\) and \(\mathbf{t}_{p_{i,j}}\) contribute significantly to this inequality. This implies that \(\mathbf{p}_{1_i}\) and \(\mathbf{p}_{1_j}\) themselves play a limited role. The reasoning is as follows: if the inequality were primarily driven by the inherent differences between \(\mathbf{p}_{1_i}\) and \(\mathbf{p}_{1_j}\), the result would hold even without the learned \(\mathbf{t}_{p_{i,i}}\) and \(\mathbf{t}_{p_{i,j}}\). However, this is clearly not the case, as ablation studies demonstrate that incorporating \(\mathbf{t}_{p}\) significantly improves retrieval accuracy.

Therefore, we posit that \(\mathbf{t}_{p_{i,i}}\), by increasing \(\text{Sim}(Y_i, X_i)\), carries semantic information less correlated with \(\mathbf{t}_{q_i}\). Consequently, it should satisfy \(s(\mathbf{t}_{q_i}, \mathbf{t}_{p_{i,i}}) \leq s(\mathbf{t}_{q_i}, \mathbf{t}_{p_{i,j}})\), ensuring that when \(\text{Sim}(Y_i, X_i) \geq \text{Sim}(Y_i, X_j)\), it follows that \(s(\mathbf{q}_{i,i},\mathbf{p}_{1_i}) \geq s(\mathbf{q}_{i,j}, \mathbf{p}_{1_j})\). It should be noted that the above discussion is based purely on empirical observations. Qualitatively, we can still visualize the representation of \( \mathbf{q}_{i,j} \) to obtain an interpretable feature space.

With more iterations, \( \mathbf{t}_{p_{i,i}} \) and \( \mathbf{t}_{p_{i,j}} \) increasingly encode semantic information from \( \mathbf{p}_{1_i} \) and \( \mathbf{p}_{1_j} \), making them more discriminative for the given query \( \mathbf{t}_{q_i} \) and reducing irregularities. Essentially, \( \mathbf{t}_{p_{i,j}} \) represents a $\mathbf{t}_{q_i}$-based version of video \( \mathbf{p}_{1_j} \), and the similarity \( s(\mathbf{t}_{p_{i,j}}, p_{1_j}) \) reflects how well $\mathbf{t}_{q_i}$ aligns with \( \mathbf{p}_{1_j} \). A higher score suggests a better alignment.

\subsubsection{Feature Space Visualization}
\begin{figure}[t]
    \centering
    \includegraphics[width=\linewidth]{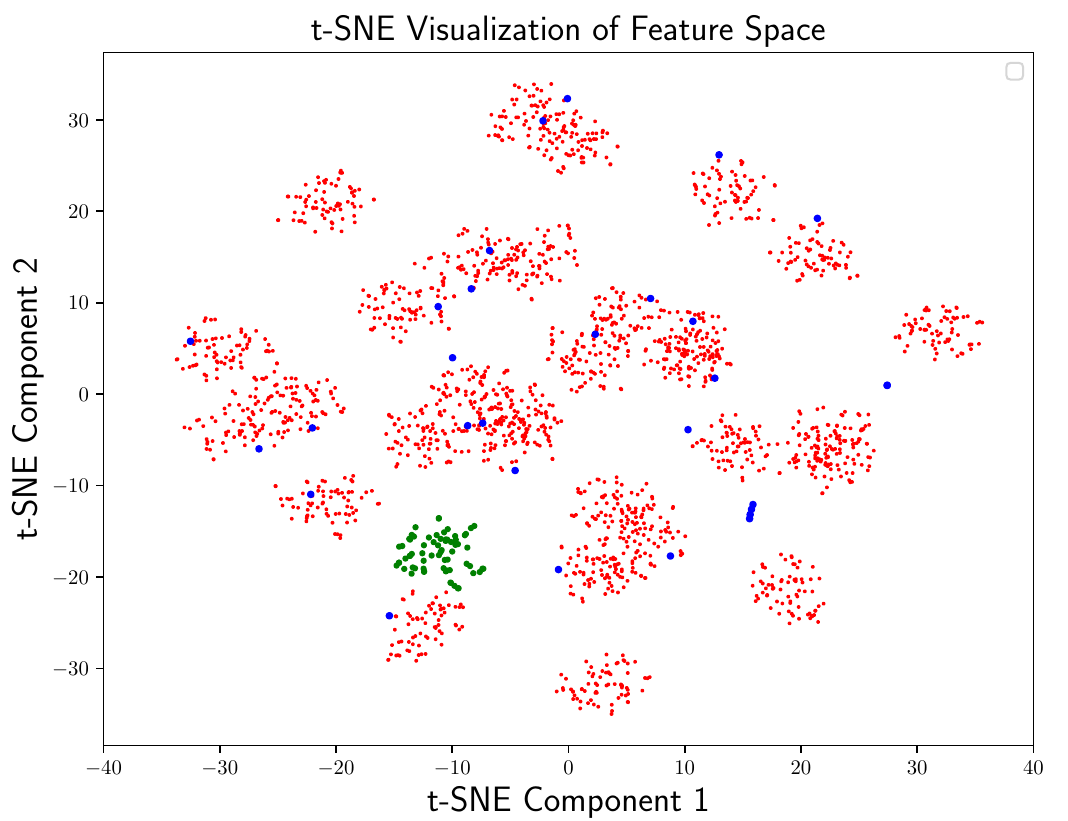}
    \caption{The t-SNE visualization of feature space, the red, blue, and green points represent \( \mathbf{q} \), video $\mathbf{p}_1$, and text $\mathbf{t}_q$, respectively.}
    \label{fig:5}
\end{figure}

In Fig.\ref{fig:5}, the red, blue, and green points represent \( \mathbf{q} \), video $\mathbf{p}_1$, and text $\mathbf{t}_{q}$, respectively. It is evident that each video representation is surrounded by numerous \( \mathbf{q} \) representations, which come from different text representations. This suggests that we do not enforce a strict alignment between text and video representations in the feature space. Instead, we approximate video representations using the proxy-derived \( \mathbf{q} \), which reduces the gap between modalities while achieving a more accurate semantic space, ultimately increasing the probability of successful retrieval.

\subsection{Complexity and Computational Cost}	
We provide the overall training procedure in Algorithm 1 and performance metrics for training and inference in Tab.\ref{tab:6}.  Since our model for text-video retrieval tasks is based on the CLIP model, it does not consume much GPU memory. Additionally, during the training phase, GPU utilization is relatively low, with most processing time dependent on the CPU. Given that we use a CPU with 48 cores, our training time is quite short. The time complexity is $O(B^2M)$. We leverage the encoder output features of the baseline network CLIP-ViP, where each video contains $M$ video [CLS] embeddings. In practice, $M$ ranges from 1 to 4, so this has a minimal impact on time complexity. The space complexity is $O(B^2)$, whereas the space complexity of CLIP-ViP is $O(B)$. This increase is due to generating $B$ proxy features for each text input, which requires greater memory usage.

\begin{table*}[ht]
\centering
\begin{tabular}{|c|c|c|c|c|c|}
\hline
\textbf{GPU}           & \textbf{Mode}      & \textbf{Time}      & \textbf{FLOPs}    & \textbf{Params}    & \textbf{Memory}      \\
\hline
2 RTX 4090             & Train             & 1h 37m             & 54.1 G            & 127.35 M           & 23071 MB             \\
1 RTX 4090             & Inference         & 6.4 s              & -                 & -                  & 4391 MB              \\
\hline
\end{tabular}
\caption{Performance metrics for training and inference}
\label{tab:6}
\end{table*}

\begin{algorithm}[tb]
\caption{Text Proxy Generation}
\label{alg:algorithm}
\textbf{Input}: Training data (text-video pairs) and hyperparameters; initialized model\\
\textbf{Output}: Trained model
\begin{algorithmic}[1] 
\FOR{$i \leftarrow training\_{epochs}$}
\FOR{each batch (text batch, video batch) in $training\_data$}
\STATE $\mathbf{X} \leftarrow \phi(X), \mathbf{Y} \leftarrow \psi(Y)$ ;
\STATE Get text feature $\mathbf{t}_{q}$ from $\mathbf{Y}$, video feature $\mathbf{p}_{1}$ from $\mathbf{X}$;
\STATE Get direction leader $\mathbf{d}_{l} \leftarrow $ \text{Cross\_atten}$(\mathbf{t}_{q}, \mathbf{p}_{1}) $;
\STATE Get director $\mathbf{d} \leftarrow \delta \mathbf{t}_q - \eta \mathbf{d}_l$;
\STATE Get $M$ similarities $\mathbf{S} \leftarrow \mathbf{t}_{q}^\top\cdot [\mathbf{p}_{1}, \mathbf{p}_{2}, ..., \mathbf{p}_{M}]$;
\STATE Get dash $\mathbf{d}_s \leftarrow \text{exp}(\mathbf{SW})$  ;
\STATE Get text proxy $\mathbf{t}_{p} = \mathbf{t}_q + \mathbf{d}_s^\top \odot \frac{\mathbf{d}}{|\mathbf{d}|}$;
\STATE Evaluate $\mathcal{L}_r, \mathcal{L}_p\text{ and } \mathcal{L}_{pos}$;
\STATE Evaluate $\nabla_{\theta} \mathcal{L}$;
\STATE Update model parameters $\theta' \leftarrow \theta - lr\nabla_{\theta} \mathcal{L} $;
\ENDFOR
\STATE Compute $s(\mathbf{t}_{q}, \mathbf{p}_{1}) + \gamma s(\mathbf{t}_{p}, \mathbf{p}_{1})$ and inference;
\ENDFOR
\end{algorithmic}
\end{algorithm}

\subsection{Retrieval Case Analysis}
When textual information lacks distinctiveness, it becomes challenging for the model to retrieve the correct video. According to the proposed Modeling Irregularity problem, this difficulty arises because the model attempts to learn a general pattern across numerous sample pairs and then uses this pattern for retrieval. However, due to interference from various irregular factors, the learned pattern is not purely based on "semantic". To mitigate the influence of these factors, we aim to decompose the 1-to-$N$ relationship into $N$ 1-to-1 relationships, compelling the model to recognize the unique data pattern within each sample pair. This approach obviates the need for the model to generalize a uniform pattern and instead narrows the retrieval scope based on "semantic" dimensions, thus alleviating the model's susceptibility to irregular factors.

\subsubsection{Successful Case}

We illustrate the effectiveness of TV-ProxyNet with two example where TV-ProxyNet successfully retrieved the correct video, while the baseline failed. 

As shown in Fig.\ref{fig:6}(a), the sample “video7214” and “video7219” from MSRVTT-1k validation set \cite{xu2016msr} were analyzed. The baseline model CLIP-ViP \cite{xue2023clip} incorrectly retrieved “video7219” when given the textual query of “video7214”. It is evident that the textual descriptions of the two videos could both semantically describe the content of the other video, making it difficult to retrieve the correct video, even manually. Notably, the dynamic visual properties of the two videos differ significantly: “video7214” exhibits frequent scene changes and significant color variations, whereas “video7219” has relatively stable camera dynamics. This indicates that although the semantic information in the textual descriptions is similar, the data distributions in the video presentations are quite distinct.

Because we established a unique data pattern for any sample pair in the dataset (with a proxy representing the distinctive data pattern of a sample pair), a given text query generates a proxy that represents a unique data relationship for any video in the dataset. This results in the text query being distinct for each video. Consequently, although some textual descriptions in the dataset may be overly concise, we can establish unique data patterns between any sample pairs. If the data distributions of two videos differ significantly in non-semantic dimensions, the data distribution differences between the proxy generated by the text query and the two videos will be more pronounced. This makes the text query more discriminative specific to different videos, thereby increasing the likelihood of successful retrieval.

As illustrated in Fig.\ref{fig:6}(b), the baseline model erroneously retrieved “video8489” when queried with the text associated with “video9335”. In “video8489”, the scene portrays “a bowl of shrimp” being “held” by a person, which diverges from the action described in the query from “video9335”, namely “put ... in a hot wok”. The correct video, “video9335”, indeed contains the action “put”. We hypothesize that the baseline model's misretrieval of “video8489” stems from the dominant visual presence of shrimp in “video8489”, which persists across frames and likely led the model to overemphasize the query phrase “prawns and shrimp”, thereby neglecting the critical action “put”. By establishing unique relationships for each sample pair, our proposed method, TV-ProxyNet, mitigates the impact of non-regular factors on semantic matching. This approach enables the model to better capture the semantic significance of the action “put”, thereby facilitating more accurate retrieval decisions.

\subsubsection{Failure Case}

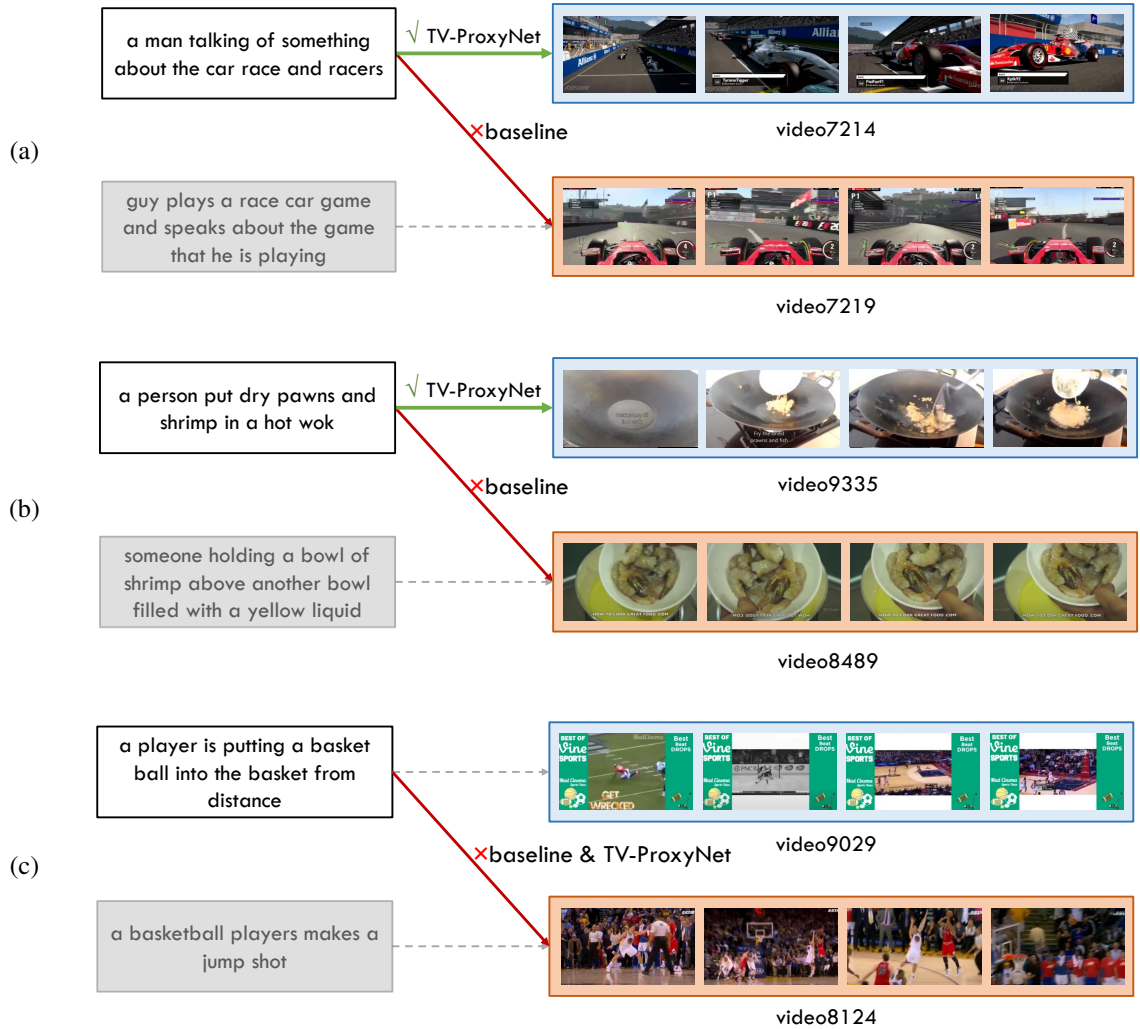
\begin{figure*}[!tbp]
    \centering

    \begin{tikzpicture}[baseline]
        \node[anchor=west] (title1) at (0,0) {(a)};
        \node[anchor=west] (image1) at (title1.east) [xshift=1em] {\includegraphics[width=0.8\textwidth]{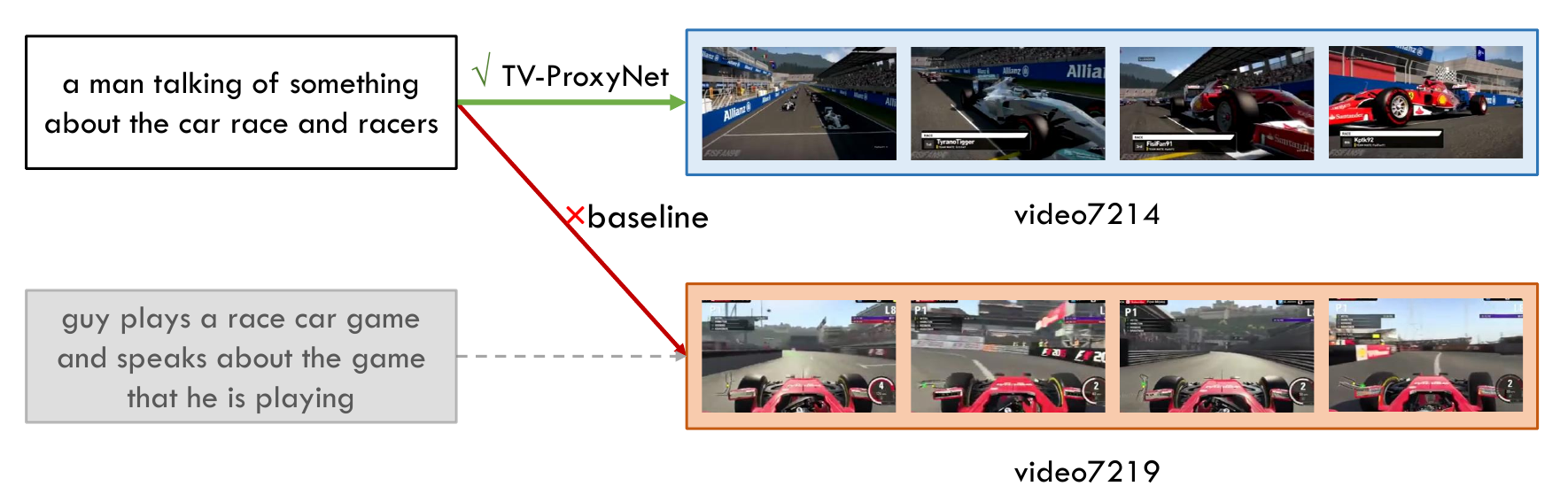}};
    \end{tikzpicture}

    \par
    \begin{tikzpicture}[baseline]
        \node[anchor=west] (title2) at (0,0) {(b)};
        \node[anchor=west] (image2) at (title2.east) [xshift=1em] {\includegraphics[width=0.8\textwidth]{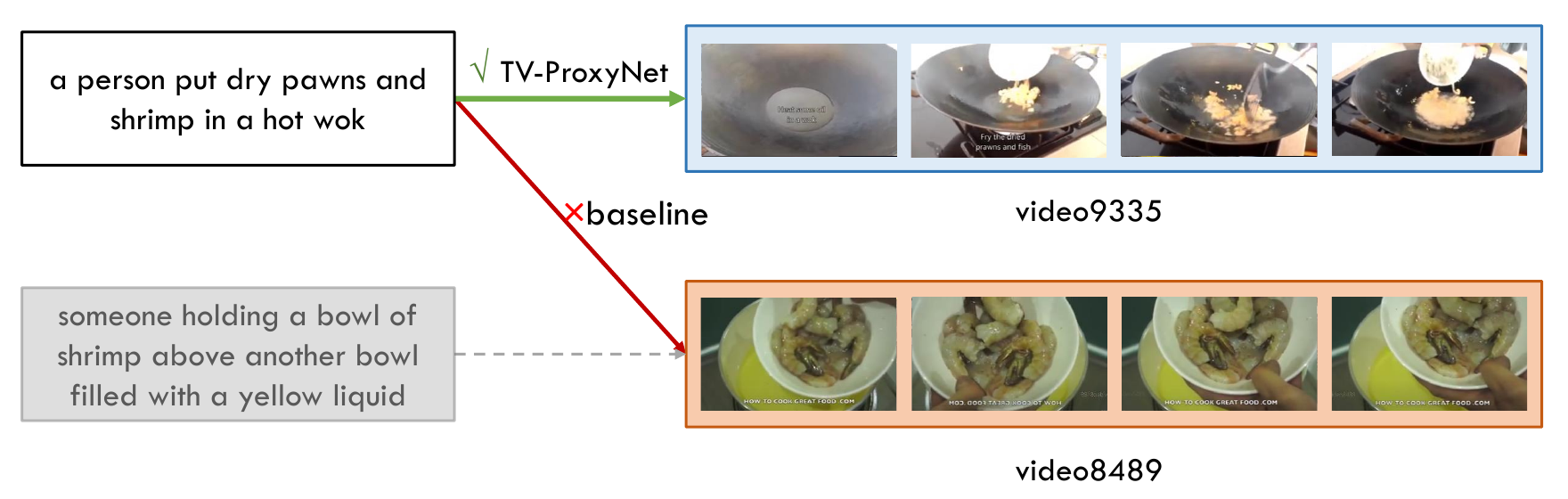}};
    \end{tikzpicture}
    
    \begin{tikzpicture}[baseline]
        \node[anchor=west] (title2) at (0,0) {(c)};
        \node[anchor=west] (image2) at (title2.east) [xshift=1em] {\includegraphics[width=0.8\textwidth]{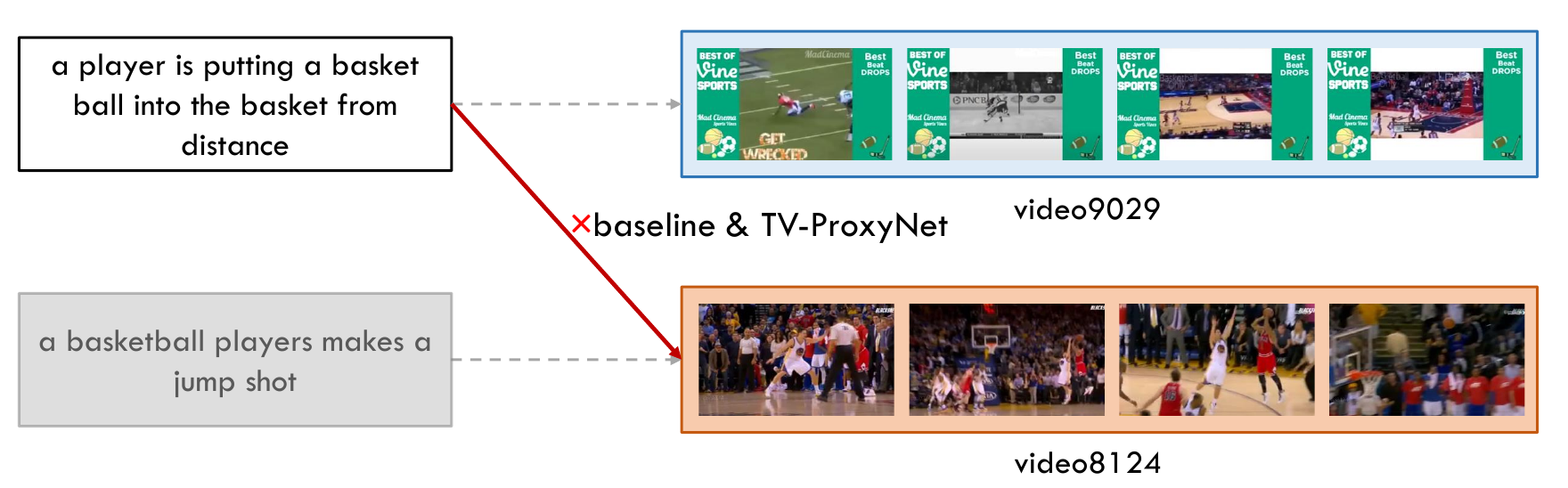}};
    \end{tikzpicture}

    \caption{Retrieval results through TV-ProxyNet and the baseline. }
    \label{fig:6}
\end{figure*}

As shown in Fig.\ref{fig:6}(c), both baseline and TV-ProxyNet erroneously retrieved “video8124” when given the textual query of “video9029”. In “video9029”, only the latter half of the video corresponds to its textual description. However, in “video8124”, the same semantic content is presented twice from different camera angles. This demonstrates that TV-ProxyNet still struggles with the problem of ambiguous textual correspondences.

